\documentclass[lettersize,journal]{IEEEtran}

\IEEEoverridecommandlockouts                              

\usepackage{color}
\usepackage{algorithmic}
\usepackage{algorithm}
\usepackage{array}
\usepackage{tabularx}
\usepackage{amsmath,amsfonts}
\usepackage{fancyhdr}
\usepackage{amssymb}
\usepackage{type1cm}
\usepackage{url}
\usepackage{caption}
\usepackage{subcaption}
\usepackage[pdftex]{graphicx}
\usepackage{bm}

\usepackage{multirow}    
\usepackage{subcaption}
\usepackage{ulem}
\usepackage{cite}
\usepackage[caption=false,font=normalsize,labelfont=sf,textfont=sf]{subfig}
\usepackage{stfloats}
\usepackage{textcomp}
\usepackage{verbatim}
\usepackage{graphicx}
\usepackage{umoline}
\usepackage{listings}

\title{\LARGE \bf
LiP-LLM: Integrating Linear Programming and dependency graph with Large Language Models for multi-robot task planning
}

\author{Kazuma Obata$^{1}$ Tatsuya Aoki$^{1}$ Takato Horii$^{1}$ Tadahiro Taniguchi$^{2}$ and Takayuki Nagai$^{1}$ 
\thanks{*This work was supported by Japan Science and Technology Agency (JST) Moonshot R\&D Grant Number JPMJMS2011.}
\thanks{$^{1}$ Dept. of Systems Innovation, Graduate School of Engineering Science, Osaka University, Osaka, Japan} 
\thanks{\tt\small \{k.obata@rlg., t.aoki@rlg., takato@, nagai@\} sys.es.osaka-u.ac.jp}
\thanks{$^{2}$ Dept. of Informatics, Kyoto University, Kyoto, Japan}
\thanks{\tt\small taniguchi@i.kyoto-u.ac.jp}
}

\begin{document}

\maketitle
\thispagestyle{empty}
\pagestyle{empty}

\begin{abstract}

This study proposes LiP-LLM: integrating linear programming and dependency graph with large language models (LLMs) for multi-robot task planning. 
In order for multiple robots to perform tasks more efficiently, it is necessary to manage the 
precedence dependencies between tasks. Although multi-robot decentralized and centralized task planners using LLMs have been proposed, none of these studies focus on 
precedence dependencies from the perspective of task efficiency or leverage traditional optimization methods.
It addresses key challenges in managing dependencies between skills and optimizing task allocation. 
LiP-LLM consists of three steps: skill list generation and dependency graph generation by LLMs, and task allocation using linear programming.
The LLMs are utilized to generate a comprehensive list of skills and to construct a dependency graph that maps the relationships and sequential constraints among these skills. 
To ensure the feasibility and efficiency of skill execution, the skill list is generated by calculated likelihood, and linear programming is used to optimally allocate tasks to each robot.
Experimental evaluations in simulated environments demonstrate that this method outperforms existing task planners, achieving higher success rates and efficiency in executing complex, multi-robot tasks. 
The results indicate the potential of combining LLMs with optimization techniques to enhance the capabilities of multi-robot systems in executing coordinated tasks accurately and efficiently.
In an environment with two robots, a maximum success rate difference of 0.82 is observed in the language instruction group with a change in the object name.

\end{abstract}

\section{Introduction}
The integration of multi-robot systems into various environments enhances the execution of a broad spectrum of tasks. 
For efficient performance, it is necessary to manage the 
precedence dependencies between tasks and identify which tasks can be executed in parallel and which cannot.
A key challenge in multi-robot task planning is managing these dependencies from natural language instructions, ensuring tasks are executed in the correct sequence, which increases the success rate.
Additionally, actions without dependencies can be parallelized, further boosting efficiency. It is also important to consider the diverse characteristics of each robot, such as the different grippers attached to the robotic arms.

Complete automation of the entire process of multi-robot task planning, including task decomposition, remains a significant challenge. Brian et al. \cite{brohan2023can} proposed SayCan, a task planning method for a single robot that uses LLMs to decompose language instructions into more detailed tasks and accomplish them. However, the relationships between decomposed tasks are simple sequence structures, and their extension to multiple robots has not been discussed. Kutter et al. \cite{kiener2010towards} modeled tasks as structures with 
precedence dependencies and proposed an approach to execute independent tasks in parallel. However, the modeling is performed manually, requiring human intervention to decompose complex tasks into simpler subtasks. 
Extensive research has been conducted on multi-robot systems \cite{rizk2019cooperative, yan2013survey}. Most multi-robot task planning systems focus on two main aspects: task decomposition and task-allocation. Task allocation is frequently automated using optimal assignment problems. 
However, as stated earlier, human involvement is necessary in many cases to ensure accurate task decomposition and handling of dependencies.

Recently, research on multi-robot task planning has explored the application of LLMs. Two primary approaches have been proposed: decentralized processing, such as RoCo \cite{mandi2024roco}, where multiple agents determine tasks through dialogue \cite{zhangbuilding, wang2024safe}, and centralized processing, such as SMART-LLMs \cite{kannan2024smart}, that use multistage reasoning to assign tasks to each agent \cite{yu2023co}. In both approaches, LLMs manage all stages of task planning, including task decomposition and task allocation.
Decentralized processing requires textual information exchange between robots, which will increase the dialogue history and expand the prompt size with the number of robots. 
Issues with hallucinations and task allocation are also common to both.
Although LLMs can be applied to various tasks, they struggle to effectively address domain-specific problems, and their performance does not match task-specific algorithmic approaches\cite{wang2024survey}. Leveraging external planners may be useful for addressing this issue.
The task allocation problem addressed in this study is classified as ST-SR-TA (Single-Task Robots, Single-Robot Tasks, Time-Extended Assignment)\cite{nunes2017taxonomy, gerkey2004formal}. Since ST-SR-TA is NP-hard, approaches have been proposed that approximate it as ST-SR-IA (Single-Task Robots, Single-Robot Tasks, Instantaneous Assignment), which can be solved using optimization, or solve it heuristically. McIntirel et al.\cite{mcintire2016iterated} proposed an approximation approach for ST-SR-IA in the context of ST-SR-TA with precedence constraints between tasks. In this study, task execution time is not considered, and the difference between the number of robots and the number of assignable tasks at each step is at most a few. Therefore, the approximation approach can achieve a solution that is sufficiently close to optimal\cite{gerkey2004formal}.

\begin{figure*}[htpb]
    \centering
     \includegraphics[width=1.0\hsize]{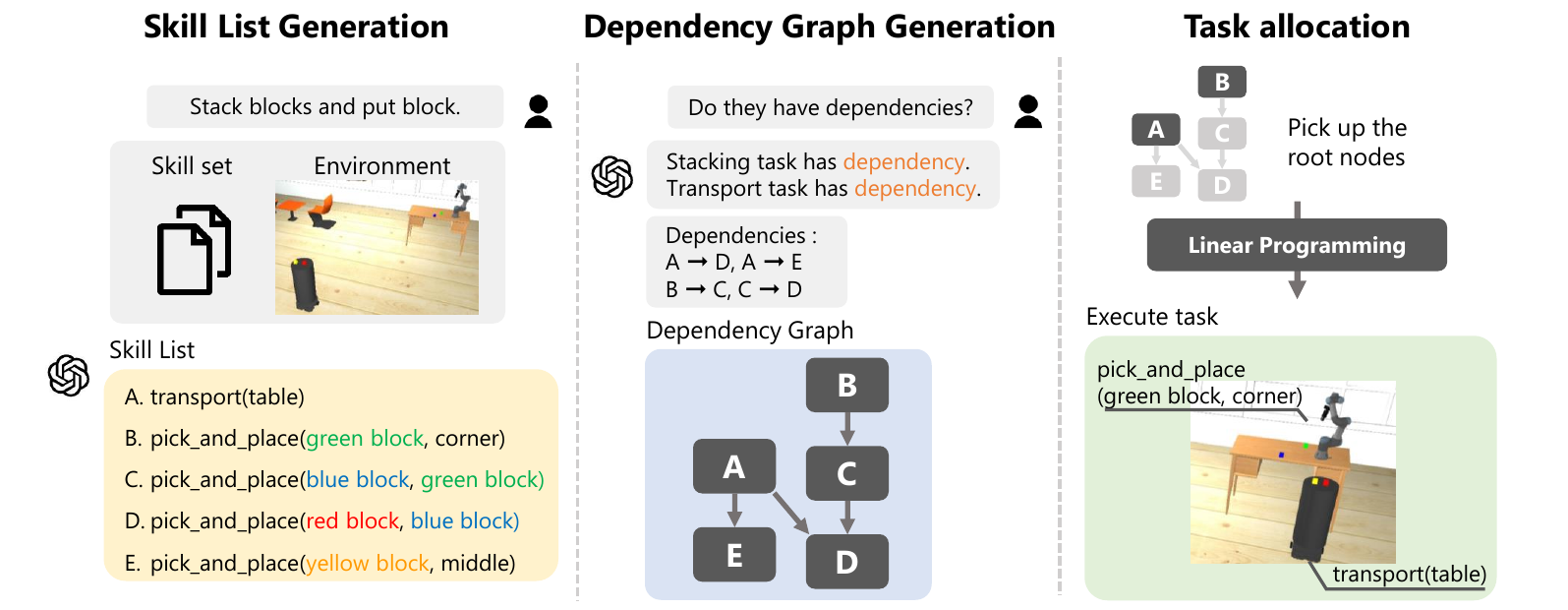}
     \caption{\textbf{LiP-LLM} generates a dependency graph from a skill list and allocates the actions to multiple robots using linear programming. In the skill list generation phase, it creates a skill list from a predefined skill set to accomplish the given instructions. Next, it analyzes the 
     skills in the skill list for 
     precedence dependencies and generates a dependency graph. Finally, it allocates the tasks to each robot using linear programming and executes them.}
     \label{fig:overview}
\end{figure*}

This study proposes a multi-robot task planning method, LiP-LLM, which combines linear programming with LLMs. As shown in Fig. \ref{fig:overview}, LiP-LLM divides task planning into three steps: skill list generation, dependency graph generation, and task allocation. LLMs manage task decomposition and dependencies traditionally handled by humans. First, LLMs generate a required skill list for a given language instruction, with SayCan’s likelihood calculation method ensuring feasible skills. To address hallucination by the LLMs, we employed the likelihood calculation used in SayCan\cite{brohan2023can} to generate the skill list. Next, a dependency graph is created, linking skills as nodes and dependencies as edges. Finally, tasks are optimally allocated to robots via linear programming, removing completed nodes from the graph after execution. This process enables effective multi-robot task allocation.

The contributions of this study are as follows
\begin{enumerate}
    \item 
    We proposed LiP-LLM to handle task dependencies by structuring the graph so that tasks with precedence dependencies are executed in the correct order.
    This allows us to structure the 
    precedence dependencies between tasks and allocate parallelizable tasks using linear programming.
    \item Multi-robot task system improves the performance over decentralized and centralized task planners proposed in related studies.
\end{enumerate}

\section{Related works}

\subsection{Application of LLMs to robotics}
Recently, LLMs have been widely applied to task planning in robotics \cite{kawaharazuka2024real}. Brian et al. \cite{brohan2023can} proposed SayCan, which is a system that integrates 
skill prediction with LLM 
and a value function to guide robotic actions.
SayCan works by combining the prediction probability from a LLM, which estimates the usefulness of a skill, with the execution probability from the value function, which assesses the likelihood that a skill can be executed correctly. When commanded, the robot detects a skill based on these combined probabilities, resulting in appropriate and feasible actions, given the language instructions.
This method leverages the extensive knowledge embedded in LLMs and supplements it with real-world information, with which that LLMs typically struggle. This integration enables generalized action generation, even for commands that contain ambiguity.

Some approaches combine LLMs with classical planners, specifically the Planning Domain Definition Language (PDDL), to perform a variety of planning tasks \cite{liu2023llm+, guan2023leveraging, zhou2024isr, capitanelli2023framework}. These studies demonstrate that converting natural language commands into PDDL using LLMs and then using a conventional planner could provide solutions to a wide range of problems.
In their work on using LLMs for robot task representation, Cao et al. \cite{cao2023robot} utilized pre-trained LLMs to generate behavior-tree-based tasks for robots. Similarly, Lin et al. \cite{pmlr-v235-lin24k} proposed the "Plan Like a Graph," which calculated the optimal execution time of tasks by transforming a sequence of tasks with ordering constraints into a graph structure using LLMs.
Additionally, efforts have been made to leverage LLMs to create robot application programming interface and other behavioral programs \cite{vemprala2024chatgpt, singh2023progprompt, liang2023code}, as well as to determine robot behavior \cite{jiang2023vima, kawaharazuka2024cooking, huang2023inner} and improve the accuracy of robotic actions\cite{pmlr-v229-zitkovich23a}.
As these studies indicated, the LLM can be effectively applied to robotics, suggesting their potential for enhancing robot task planning through human robot interaction.

\subsection{Utilization of LLMs for multi-robot systems}

Zhao et al. \cite{mandi2024roco} proposed RoCo, a system in which multiple robots interactively discuss task strategies using LLMs to perform tasks and trajectory planning. RoCo leverages the role-setting and interactive nature of ChatGPT by assigning names to each robot and providing environmental feedback, resulting in a highly interpretable and flexible system. This approach demonstrated a high success rate in six benchmark tasks that emphasized the sequential and overlapping nature of a robot's workspace. 
However, the benchmarks involved a relatively small number of robots (2-3), and action decisions were made in a constrained environment with limited action options for each robot. The scalability of the number of robots and limitations on the number of actions were not addressed.
Shyam et al. \cite{kannan2024smart} proposed SMART-LLM, a task planning framework for multiple robots using LLMs. This method involves task decomposition, team formation, and task assignment based on input language instructions, subdividing tasks and assigning them according to robot characteristics. 
However, the task decomposition granularity in this method is coarse, and the framework was not evaluated for instructions requiring complex task decomposition.

Chen et al. \cite{chen2024scalable} evaluated agent structures and prompt configurations for a multi-robot system using LLMs. They proposed and compared four types of structures: conventionally distributed structure and two centralized hybrid types. The study found that one of the hybrid types achieved a higher performance.
Chen et al. \cite{chen2024autotamp} proposed Autotamp, an approach that uses LLMs and STL for multi-robot planning, but it is limited to mobile robots and does not address the problem settings that include manipulators.
Zhang et al. \cite{zhangbuilding} introduced a method for determining actions through natural language communication using LLMs in a multi-agent system. Yu et al. \cite{yu2023co} developed Co-NavGPT, a cooperative navigation system for multiple robots using LLMs. In this system, environmental information, such as obstacles and uncharted areas on a map, is converted into text, thereby enabling cooperative navigation.
Furthermore, research on the application of LLMs for multi-agent systems extends beyond robotics, with various studies exploring their applications in different domains, such as virtual game worlds and education\cite{ijcai2024p890}.

Thus, various approaches for multi-robot task planning using LLMs were proposed. However, these studies did not comprehensively compare the different methods or adequately evaluate their performances.
In this study, a series of task planning with natural language as input is realized in a multi-robot task plan by having LLMs handle task dependencies that are previously managed by humans.

\section{Proposed Method}

\subsection{Skill List Generation}
\label{section:アクションリスト生成}

The algorithmic procedure for skill list generation is outlined in Algorithm \ref{alg:saycan}.
In the skill list generation process, we used LLMs to generate a list of skills necessary to accomplish given language instructions from a predefined skill set. First, a set of executable robot skills $\Pi$ was defined. 
Each 
skill language description $s_{\pi} \in s_{\Pi}$ was represented by a command in the format of function + arguments, such as "pick\_and\_place(red block, middle)", which a robot can execute.
The probability that skill 
$s_{\pi}$ is predicted for language instruction $i$ is represented as 
$p(s_{\pi|i})$ and can be computed using the LLM. 
In lines 4-5, for
each skill 
$s_{\pi} \in s_{\Pi}$, $p(s_{\pi}|i)$ is calculated, and skill $s_{n}$ with the highest probability 
is selected 
in step $n$ from $C$, which represents the set of predicted probabilities for each skill(shown in lines 3, 6-8). Subsequently, skill $s_{n}$ is appended to the end of language instruction $i$, and the next skill is selected 
in lines 1, 8-9. This process is repeated until the end command is selected by 
$p(s_{\pi}|i, s_{n-1}, \dots , s_{0})$(lines 2-10).
By using the predicted probabilities of skills instead of directly generating text, we can prevent hallucinations where undefined skills are output.

The prompt configuration used in this step is shown below.
\begin{enumerate}
    \item \textbf{Purpose of inference and brief description}: 
    Describe the number of robots present in the environment and the objective of the inference.
    \item \textbf{Rules}: 
    Describe basic rules, including explanations of what each generated skill represents.
    \item \textbf{Considerations for Inference}: 
    Provide information to consider during inference, such as the necessary conditions for object transfer.
    \item \textbf{Few-shot Examples \cite{brown2020language}}: 
    The Few-shot method is used to improve the prediction performance. For this method, five examples are provided.
\end{enumerate}

\begin{figure}[!t]
\begin{algorithm}[H]
    \caption{Skill List Generation}
    \label{alg:saycan}
    \begin{algorithmic}[1]
    \renewcommand{\algorithmicrequire}{\textbf{Input:}}
    \renewcommand{\algorithmicensure}{\textbf{Output:}}
    \REQUIRE Language instruction $i$, predefined skill set $\Pi$

    \STATE $n = 0
    $

    \WHILE{$
    s_{\pi_{n-1}} \neq "done"$}
    
    \STATE $C = \emptyset$
    
    \FOR{$
    s_{\pi} \in s_{\Pi}$}
    
    \STATE $
    p(s_{\pi} | i, s_{n-1}, \ldots , s_{0})\leftarrow $LLMs$(prompt + i + s_{0} + \ldots + 
    s_n)$
    
    \STATE $C = C \cup 
    p(s_{\pi} | i, s_{n-1}, \ldots , s_{0})$
    
    \ENDFOR
    
    \STATE $s_{n} = 
    \rm{argmax}_{s_{\pi} \in s_{\Pi}}C$
    
    \STATE $n = n + 1$
    
    \ENDWHILE
    
    \ENSURE $S$: skill list
    \end{algorithmic}
\end{algorithm}
\end{figure}

\subsection{Dependency Graph Generation}
\label{section:グラフ構造化}



Taking as input the skill list $S$ and language instruction generated in the previous steps, 
Directed Acyclic Graph (DAG) $ G = (V, E) $ is generated based on the dependencies using LLMs. In this graph:
\begin{itemize}
    \item Each node $v_i \in V$ represents a skill, where $v_i$ corresponds to each element in a sequence of skills from the previous steps.
    \item Each edge $e_{ij} \in E$ represents the dependencies between skills, indicating that skill $v_j$ depends on skill $v_i$.
\end{itemize}

This process involves leveraging LLMs to infer dependencies and construct a dependency graph that captures the sequential relationships among skills, thus enabling parallelizable skill selection. In this study, a "dependency" is defined as the relationship between two tasks where task B cannot be executed unless task A has been completed. This dependency ensures that tasks are performed in the correct order.

We utilized a step-wise inference method known as Chain of Thought
 \cite{wei2022chain} to enhance the inference performance of the language model during dependency graph generation. 
First, we generated the dependency relationships between skills in textual form. Next, we generated $T_{S}$, which represents the edges between nodes in the text format "${v_i} \rightarrow {v_j}$"
in line 2 of Algorithm \ref{alg:allocation}. 
In line 3, this
output was 
converted into edge $e_{ij} \in E_{S}$ to create graph $G_{S}$
$(V_S, E_S)$, consisting of nodes $V_S$, representing each element of the skill list, and edges $E_S$.
In line 1-4, we detect whether the generated graph contains cycles, and if a cycle is found, the dependencies are regenerated.
As described in the previous section, the prompt includes information regarding the surrounding environment of the robot to provide a comprehensive context.

\subsection{Task Allocation}

This allocation was achieved using an optimal assignment problem approach, which is commonly employed in task allocation for multi-robot systems.
The algorithm of Dependency Graph Generation and Task Allocation is presented in Algorithm \ref{alg:allocation}.

\begin{enumerate}
    \item {\bf Election of skills with no dependencies}\\
    A dependency graph is a directed graph with skills as nodes and dependencies between skills as edges. 
    We searched for the root nodes $V_{r}$ within the graph and selected them as executable skills
    (lines 6-11). As the selected nodes did not have parent nodes, no dependencies were existed between them, enabling parallel execution.
    \item {\bf Weight Calculation}\\
    The weights for each robot's skills were calculated. This experiment involved two robot types: an arm robot and a mobile robot. For the arm robots, weights were assigned based on skill feasibility and adjusted by distance, as operating on distant objects raises the risk of motion failure or collision with other robots. The weights were calculated based on the distance from the robot to the target object, with observation $o$ representing environmental data from the simulation(line 12).
    Weight $w_{jk}$ for each skill $k$ is calculated using the normalized value $d_{jb_{k}}'$ of the distance $d_{jb_{k}}$ between each robot $j$ and the grasping object $b_{k}$, using (\ref{eq:affordance}). $\alpha$ represents a constant ranging from 0 to 1. In this case, $\alpha$ was empirically set to 0.3.
    For mobile robots, a value of 1 or 0 is calculated as the weight, depending on the skill's executability (\ref{eq:mobile_robot_affordance}).
    
    \begin{equation}
       \label{eq:affordance}
            w_{jk}=1-\alpha d_{jb_{k}}'
    \end{equation}

    \begin{equation}
        \label{eq:mobile_robot_affordance}
            w_{jk} = 
       \begin{cases}
          1 \;\;\;\; \text{if \ executable}\\
          0 \;\;\;\; \text{otherwise}
       \end{cases}
    \end{equation}
    
    \item {\bf Optimal Assignment}\\
    To allocate tasks to multiple robots, we use a linear programming assignment problem based on skill weights. For each skill selected in Section \ref{section:アクションリスト生成}, we created an allocation problem as described in (\ref{eq:optimization}), where the sum of the skill weights for each robot is the objective function. Let $N$ and $M$ represent the number of robots and skills, respectively, and $x_{jk}$ be a variable indicating whether robot $j$ executes skill $k$.
    
    \begin{equation}
       \begin{aligned}
        \centering
        & \text{maximize} & z= \sum_{j}^{N} \sum_{k}^{M} w_{jk}x_{jk}  \hspace{3em}\\
        &\text{subject to} &\sum_{j}^N x_{jk} \leq 1, \sum_{k}^M x_{jk} \leq 1,\hspace{1.8em}\\
            & &x_{jk} \in \{0, 1\}, \forall{j}\in N, \forall{k}\in M\hspace{1em}\\
       \end{aligned}
       \label{eq:optimization}
    \end{equation}
    
    Robot can simultaneously perform at most one skill, and at most one robot can be assigned to each skill.
    Therefore, we imposed the constraint that each robot was assigned no more than one skill, and that each skill was assigned to no more than one robot.
    
    In lines 13-14, the
    solution to this allocation problem assigned skill $V_{e}$ to each robot. Any robot that is not assigned a skill is given the waiting skill "stay()". Once a robot completed its skill and provided feedback, the corresponding node $V_{e}$ and its associated edges were removed from dependency graph $G_{S}$
    (line 15).
    These steps were repeated until all nodes in the graph were removed or until no more skills can be performed
    (lines 5-16).
\end{enumerate}



\begin{figure}[!t]
\begin{algorithm}[H]
    \caption{Dependency Graph Generation and Task Allocation}
    \label{alg:allocation}
    \begin{algorithmic}[1]
    \renewcommand{\algorithmicrequire}{\textbf{Input:}}
    \renewcommand{\algorithmicensure}{\textbf{Output:}}
    \REQUIRE Language instruction $i$, observation $o$, skill list $S$

    \WHILE{CycleDetection(Gs) $= True$}
    
    \STATE $T_{S} \leftarrow $LLMs$(prompt + i + S$)
    \STATE $G_{S} 
    (V_{S}, E_{S})\leftarrow $CreateGraph$(S,
    T_{S})$

    \ENDWHILE

    \WHILE{$
    V_{S}\neq \emptyset$}
    \STATE $V_{r} = \emptyset$

    \FOR{$v \in 
    V_{S}$}
    
    \IF{$v \ is \ $rootnode}
    \STATE $V_{r} = V_{r} \cup v$
    \ENDIF
    \ENDFOR

    \STATE $W_{jk} \leftarrow $CaluculateWeights$(o)$
    \STATE $V_{e} \leftarrow $OptimizationSolver$(V_{r}, W_{jk})$

    \STATE Execute $V_{e}$
    \STATE $G_{S} \leftarrow $DeleteNode$(G_{S}, V_{e})$
    
    \ENDWHILE
    \end{algorithmic}
\end{algorithm}
\end{figure}
\section{Experiment}

Two types of experiments were conducted to evaluate the performance of the proposed LiP-LLM.
Experiment 1 evaluated the versatility of the planner by examining whether it could appropriately plan various instructions. In Experiment 2, experiments were conducted in a more complex environment with a larger number of robots than in Experiment 1, to evaluate the limitations of the planner's performance and scalability.

In this experiment, the AWS RoboMaker Small House World ROS package\footnote{https://github.com/aws-robotics/aws-robomaker-small-house-world} was used to simulate a home environment. The furniture arrangement was partially modified for this experiment. 
Four colored blocks and bowls (blue, red, yellow, and green) were placed in the experimental environment.
The robot was assumed to know the locations of the objects. Three experimental environments were used with variation in the type and number of robots.
LiP-LLM used OpenAI's GPT model "text-davinci-003" as the LLM.

\begin{figure*}[hbtp]
    \centering
    \begin{minipage}[b]{0.30\linewidth}
        \centering
        \includegraphics[keepaspectratio, height=2.5cm]{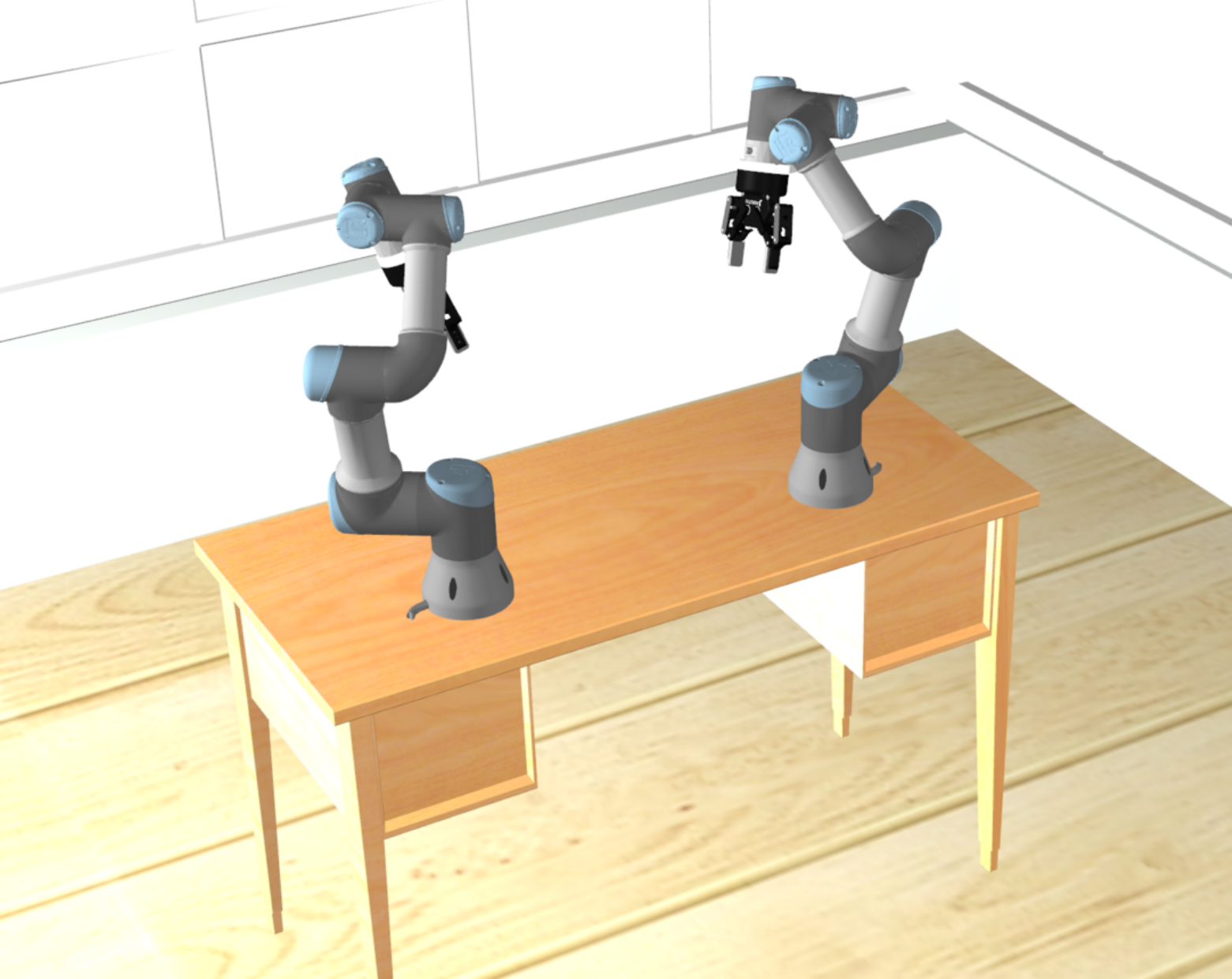}
        \subcaption{Environment A}
        \label{fig:envA}
    \end{minipage}
    \begin{minipage}[b]{0.30\linewidth}
        \centering
        \includegraphics[keepaspectratio, height=2.5cm]{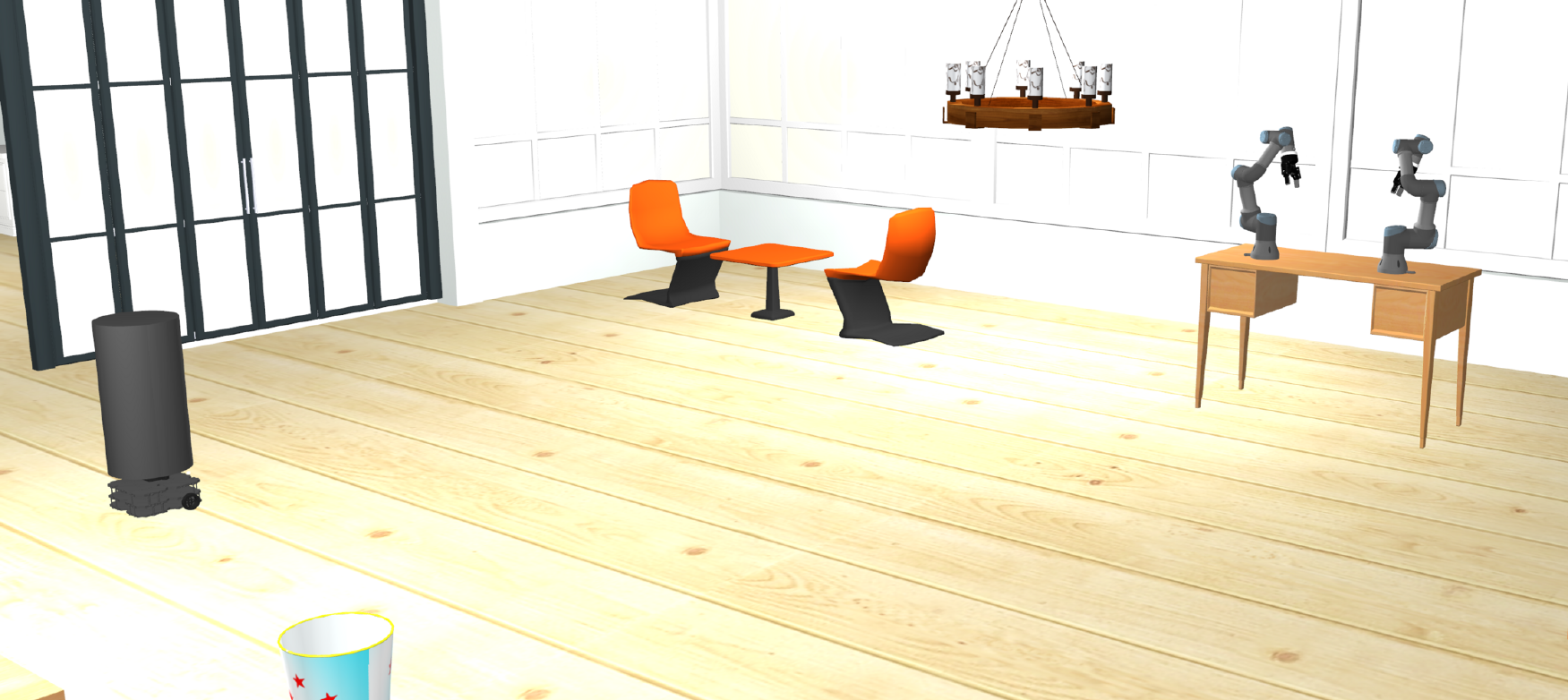}
        \subcaption{Environment B}
        \label{fig:envB}
    \end{minipage}
    \begin{minipage}[b]{0.30\linewidth}
        \centering
        \includegraphics[keepaspectratio, height=2.5cm]{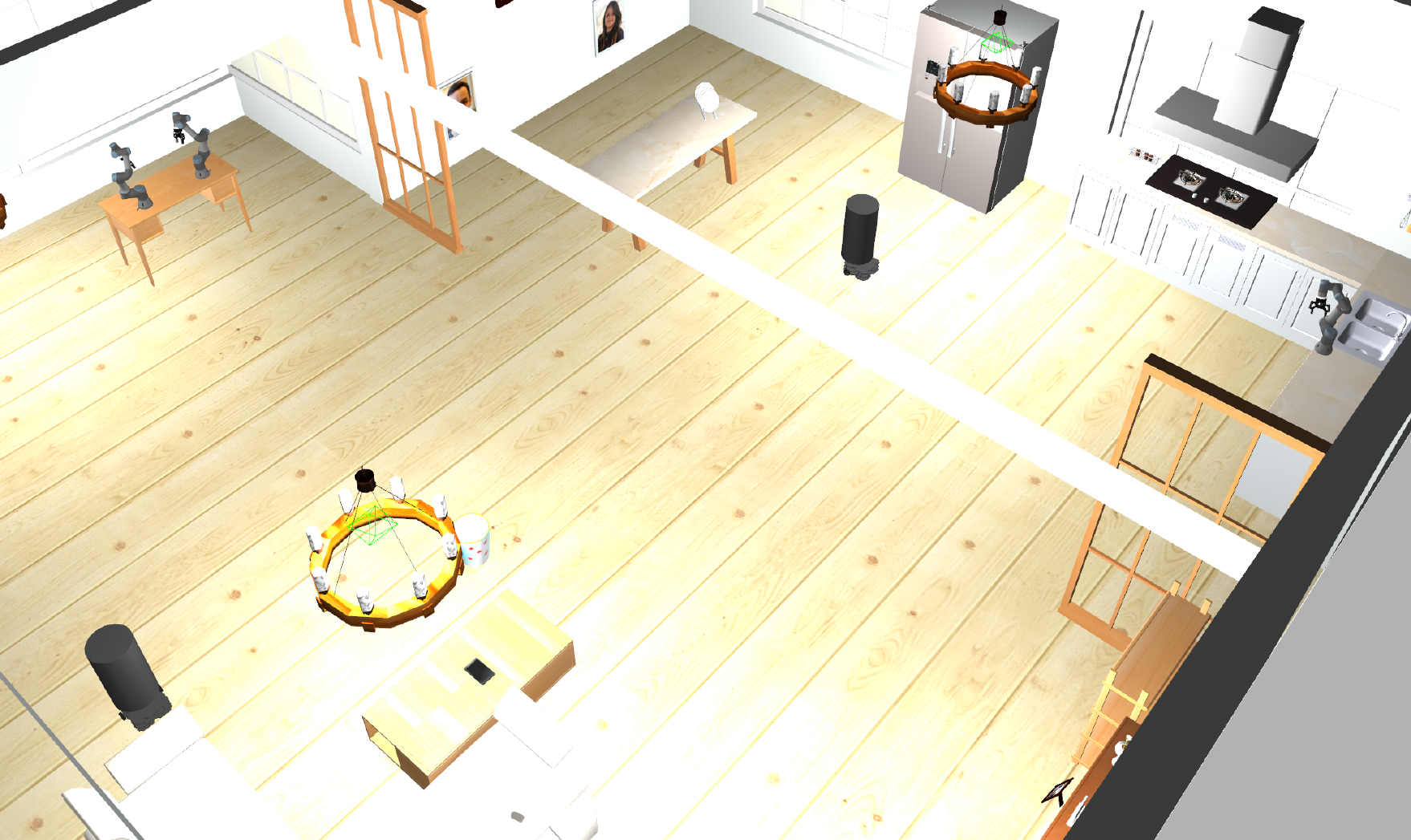}
        \subcaption{Environment C}
        \label{fig:envC}
    \end{minipage}
    \caption{Experiment environments :
    (a)Environment A includes two arm robots. 
    (b)Environment B includes two arm robots and one mobile robot. 
    (c)Environment C includes three arm robots (two in the living room and one in the kitchen) and two mobile robots.}
\end{figure*}

\begin{itemize}
    \item \textbf{Environment A}: 
    This setup featured a living room with two arm robots (Fig. \ref{fig:envA}). These robots can interact with object placement and stacking tasks on a table.
    \item \textbf{Environment B}: 
    This environment included a living room with two arm robots and a mobile robot (Fig. \ref{fig:envB}). The robots performed object placement and stacking tasks on a table, with some objects requiring transportation by the mobile robot.
    \item \textbf{Environment C}: 
    This configuration comprised two arm robots in the living room, one arm robot in the kitchen, and two mobile robots (Fig. \ref{fig:envC}). The robots performed tasks either at the table or in the kitchen and might need to move objects between these areas.
\end{itemize}

\subsection{Task Setting}

\subsubsection{Experiment 1}

In Experiment 1, we evaluated the generality of LiP-LLM by changing two variables: language instructions and constraints provided to the robot in Environments A and B.

\begin{enumerate}
    \item {\bf Instruction}: 
    Four groups of instructions were used to evaluate generality, resulting in 63 different languages instructions. The list of instructions is shown in TABLE \ref{tab:instruction}.

\newcolumntype{C}{>{\centering\arraybackslash}X}

\begin{table*}[]
    \centering
    \caption{Language Instruction\\ 
    'Group Name' represents the instruction group. 'Detail' provides details about each instruction group. 'Example' shows example of the instructions. 'Trial' is the number of instructions in each group.}
    \label{tab:instruction}
    \renewcommand{\arraystretch}{1.2}
    
    \scalebox{0.9}{
    \begin{tabularx}{1.1\linewidth}{|C|p{28.9em}|p{26.6em}|p{1.8em}|}
        \hline
        Group name & \hspace{13em}Detail & \hspace{12em}Example & 
        Trial\\
        \hline \hline
        Group1:\par Basic & Instruction composed of a stacking task and a placement task, which are the basic operations among the tasks assumed in this study. & Stack the blocks in the order of red, yellow on the middle, and put green block and blue block on the different corner of the table. & \vspace{-0.3em}\hspace{0.3em}15\\
        \hline
        Group2:\par Name Change & Instruction refers to a name that differs from that of the object or location defined in the skill, such as "block" being described as "box". & Put green cube on the lower right edge, and stack cubes in order of red, blue on the upper right edge. & \vspace{-0.3em}\hspace{0.3em}15\\
        \hline
        Group3:\par Ambiguity & Name and location of the block are euphemistic and ambiguous, such as "place the block clockwise. & Put the blocks in the order of red, blue, yellow, and green clockwise starting from the top right corner. & \vspace{-0.3em}\hspace{0.3em}15\\
        \hline
        Group4:\par Add Bowl & Instruction consisting of an environment with an additional bowl in the environment. & Put bowl to the middle, then put red, green, and blue block in the bowl. & \vspace{-0.25em}\hspace{0.3em}18\\
        \hline
    \end{tabularx}
    }
    
\end{table*}

    \item {\bf Constraints on the robot}: 
    Each robot in an environment has different characteristics. Two arm robots in living rooms have constraints on the objects they can grasp.
    \begin{enumerate}
        \item Condition 1: 
        There are no restrictions on the robots, allowing them to grasp any object in the environment.
        \item Condition 2: 
        Robot 1 can grasp red, blue, and yellow objects (blocks and bowls), whereas Robot 2 can only grasp red and green objects.
        \item Condition 3: 
        Robot 1 can grasp red and blue objects, and Robot 2 can grasp yellow and green objects.
    \end{enumerate}
\end{enumerate}

\subsubsection{Experiment 2}

We evaluated the utility of LiP-LLM by varying three variables: language instructions, constraints imposed on robots, and object placement positions in Environment C.

\begin{enumerate}
    \item {\bf Instruction}: 
    The same 15 language instructions from the instruction group (basic instructions used in Experiment 1) were applied, with the same structure but different installation positions.
    \item {\bf Constraints on the robot}: 
    The arm robot to be placed in the living room were subjected to the same constraints as in Experiment 1. However, no constraints are imposed on the arm robot placed in the kitchen.
    \item {\bf Location of objects}: 
    In Environment C, task planning results can vary significantly based on the object placement, even when using the same language instructions. Therefore, three difficulty levels were set based on the necessity of object transportation by the mobile robot:

    \begin{enumerate}
        \item \textbf{Level 1}: 
        Objects were placed close to the target positions,thus requiring no object transportation for the initial placement. This is the least challenging setting as the mobile robot does not need to transport objects.
        \item \textbf{Level 2}: 
        Objects may be held by the mobile robot instead of leaving them at the target position, which requires initial transportation by the mobile robot. This setting has a medium difficulty level because the mobile robot needs to transport objects.
        \item \textbf{Level 3}: 
        Objects are placed in a room different from the target location, necessitating object transportation between rooms. This is the most challenging setting, requiring coordination between the arm and mobile robots to transport the objects.
    \end{enumerate}
\end{enumerate}

\subsection{Comparative methods}

\begin{itemize}
    \item \textbf{RoCo} \cite{mandi2024roco}: 
    A method in which multiple robots interactively make skilled decisions using LLMs. The LLMs uses OpenAI's GPT-4-0613.
    \item \textbf{SMART-LLM} \cite{kannan2024smart}: 
    A task planning method in which 
    a series of operations, including 
    task decomposition, 
    team formation, and task allocation were performed using LLMs.
    The LLMs also used the OpenAI's GPT-4-0613.
\end{itemize}

\subsection{Evaluation index}


\begin{enumerate}
    \item {\bf Success Rate}: 
    The success rate was recorded as 1 if a plan that fulfilled a language instruction was generated and 0 if it failed. Only trials with a success rate of one were considered in the computation time metrics.
    \item {\bf Success weighted by Path Length(SPL) \cite{anderson2018evaluation}}: 
    This metric evaluates plan efficiency by comparing the ideal shortest number of steps, determined by a human, with the steps in the generated plan. In this study, the SPL reflects success rates weighted by step count rather than path length.
    In this experiment, (\ref{eq:spl}) was used.
    \begin{equation}
        \frac{1}{N} \sum^{N}_{i=1} S_{i} \frac{l_{i}}{max(p_{i}, l_{i})}
        \label{eq:spl}
    \end{equation}
    where $N$ is the number of trials, $S_{i}$ is one for success and zero for failure, $l_{i}$ is the minimum number of steps in trial $i$, and $p_{i}$ is the number of steps in trial $i$.
    \item {\bf Process Time}: 
    The time taken by the planner to process the allocation and planning from the time it receives instructions to the time it sends commands to the robots.
\end{enumerate}

\subsection{Result}

\subsubsection{Experiment 1}

\newcolumntype{C}{>{\centering\arraybackslash}X}
\newcolumntype{D}{>{\centering}p{5.2em}}
\newcolumntype{E}{>{\centering}p{3em}}

\begin{table}[tb]
    \caption{\\Experiment 1: Results for each instruction group}
    \label{table:exp1_group}
    \centering
    \renewcommand{\arraystretch}{1.2}

    \begin{tabularx}{\linewidth}{|p{0.3em}|D||E|C|C|E|C|C|} \hline
             & \multirow{2}{*}{\vspace{-1.1em}methods} & \multicolumn{3}{|c|}{Environment A} &\multicolumn{3}{|c|}{Environment B} \\ \cline{3-8}
             & ~ &Success Rate$\uparrow$&\vspace{-0.25em}SPL$\uparrow$&Time [s]$\downarrow$& Success Rate$\uparrow$ &\vspace{-0.25em}SPL$\uparrow$&Time [s]$\downarrow$ \\\hline \hline
            \multirow{3}{*}{\hspace{-0.5em}G1} & LiP-LLM & \uuline{$\bm{0.93}$} & \uuline{$\bm{0.89}$} & \uuline{$\bm{21.0}$} & \uuline{$\bm{0.82}$} & \uuline{$\bm{0.72}$} & \uuline{$\bm{33.2}$} \\
            ~ & RoCo & \uline{0.71} & \uline{0.60} & \uline{26.3} & \uline{0.58} & 0.44 & 86.9 \\
            ~ &\hspace{-0.54em}SMART-LLM& 0.62 & 0.59 & 33.4 & 0.49 & \uline{0.45} & \uline{49.4} \\\hline\hline
            \multirow{3}{*}{\hspace{-0.5em}G2} & LiP-LLM & \uuline{$\bm{0.93}$} & \uuline{$\bm{0.88}$} & \uuline{$\bm{21.9}$} & \uuline{$\bm{0.91}$} & \uuline{$\bm{0.81}$} & \uuline{$\bm{35.5}$} \\
            ~ & RoCo & \uline{0.69} & \uline{0.45} & 47.9 & \uline{0.36} & \uline{0.22} & 117 \\
            ~ &\hspace{-0.54em}SMART-LLM& 0.11 & 0.11 & \uline{46.0} & 0.22 & 0.19 & \uline{66.0} \\\hline\hline
             \multirow{3}{*}{\hspace{-0.5em}G3} & LiP-LLM & \uuline{$\bm{0.44}$} & \uuline{$\bm{0.40}$} & \uuline{$\bm{25.0}$} & \uuline{$\bm{0.44}$} & \uuline{$\bm{0.43}$} & \uuline{$\bm{37.2}$} \\
            ~ & RoCo & \uuline{$\bm{0.44}$} & \uline{0.37} & 46.6 & 0.18 & 0.12 & 105 \\
            ~ &\hspace{-0.54em}SMART-LLM& 0.36 & 0.34 & 45.3 & \uline{0.29} & \uline{0.17} & \uline{60.9} \\\hline\hline
             \multirow{3}{*}{\hspace{-0.5em}G4} & LiP-LLM & \uuline{$\bm{0.50}$} & \uuline{$\bm{0.46}$} & \uuline{$\bm{25.4}$} & \uuline{$\bm{0.17}$} & \uuline{$\bm{0.15}$} & \uuline{$\bm{40.6}$} \\
            ~ & RoCo & 0.22 & 0.18 & 66.0 & 0.09 & 0.06 & 117 \\
            ~ &\hspace{-0.54em}SMART-LLM& \uline{0.37} & \uline{0.37} & \uline{48.8} & \uline{0.13} & \uline{0.11} & \uline{52.6} \\\hline
    \end{tabularx}

\end{table}

\begin{table}[tb]
    \caption{\\Experiment 1: Result for each constraint}
    \centering
    \renewcommand{\arraystretch}{1.2}
    
    \begin{tabularx}{\linewidth}{|p{0.4em}|D||E|C|C|E|C|C|} \hline
         & \multirow{2}{*}{\vspace{-1.1em}methods} & \multicolumn{3}{|c|}{Environment A} &\multicolumn{3}{|c|}{Environment B} \\ \cline{3-8}
         & ~ &Success Rate$\uparrow$&\vspace{-0.25em}SPL$\uparrow$&Time [s]$\downarrow$& Success Rate$\uparrow$ &\vspace{-0.25em}SPL$\uparrow$&Time [s]$\downarrow$ \\\hline \hline
        \multirow{3}{*}{\hspace{-0.5em}C1} & LiP-LLM & \uuline{$\bm{0.70}$} & \uuline{$\bm{0.67}$} & \uuline{$\bm{22.8}$} & \uuline{$\bm{0.56}$} & \uuline{$\bm{0.51}$} & \uuline{$\bm{34.3}$} \\
        ~ & RoCo & \uline{0.55} & 0.43 & \uline{37.5} & 0.25 & 0.17 & 97.7 \\
        ~ & \hspace{-0.54em}SMART-LLM & 0.48 & \uline{0.46} & 40.9 & \uline{0.33} & \uline{0.29} & \uline{58.1} \\\hline\hline
        \multirow{3}{*}{\hspace{-0.5em}C2} & LiP-LLM & \uuline{$\bm{0.67}$} & \uuline{$\bm{0.60}$} & \uuline{$\bm{22.7}$} & \uuline{$\bm{0.60}$} & \uuline{$\bm{0.56}$} & \uuline{$\bm{36.2}$} \\
        ~ & RoCo & \uline{0.49} & \uline{0.36} & 53.2 & \uline{0.33} & \uline{0.23} & 96.9 \\
        ~ & \hspace{-0.54em}SMART-LLM & 0.30 & 0.29 & \uline{42.3} & 0.24 & 0.18 & \uline{57.1} \\\hline\hline
         \multirow{3}{*}{\hspace{-0.5em}C3} & LiP-LLM & \uuline{$\bm{0.71}$} & \uuline{$\bm{0.67}$} & \uuline{$\bm{22.8}$} & \uuline{$\bm{0.52}$} & \uuline{$\bm{0.47}$} & \uuline{$\bm{34.2}$} \\
        ~ & RoCo & \uline{0.49} & \uline{0.39} & 53.2 & \uline{0.29} & 0.20 & 109 \\
        ~ & \hspace{-0.54em}SMART-LLM & 0.32 & 0.31 & \uline{41.8} & 0.25 & \uline{0.21} & \uline{51.9} \\\hline
    \end{tabularx}
    \label{table:exp1_condition}
\end{table}

    \begin{enumerate}
        \item Success Rate:
        As shown in TABLEs \ref{table:exp1_group} and \ref{table:exp1_condition}, LiP-LLM achieved the highest success rate in Environments A and B, with a maximum difference of 0.82. TABLE \ref{table:exp1_group} indicates that LiP-LLM maintains robust task planning even when language commands aim to induce failures. TABLE \ref{table:exp1_condition} shows no significant changes across methods under varying constraints.
        \item SPL: 
        The SPL results followed a similar trend to the success rate, indicating efficiency differences. SMART-LLM, LiP-LLM, and RoCo scored higher in efficiency, with a maximum difference of 0.77 compared to the baseline. Although SMART-LLM had a lower success rate, its SPL was not high because in successful trials, it generated near-optimal plans. LiP-LLM and RoCo were slightly less efficient due to redundant tasks.
        \item Process Time: 
        LiP-LLM outperformed SMART-LLM and RoCo in terms of process time. This showed a difference of up to approximately 80 s from the baseline method. 
        In the proposed method, the waiting time for the LLM invocation accounted for the majority of the processing time. Therefore, the processing time increased with the number of inference calls in both the proposed and comparative methods.
    \end{enumerate}

\subsubsection{Experiment 2}

    \begin{enumerate}
        \item Success Rate: 
        As shown in TABLEs \ref{table:exp2_level} and \ref{table:exp2_condition}, LiP-LLM had the highest success rate, with a maximum difference of 0.34, similar to Experiment 1. Although success rates were lower for all methods in the more complex long-term tasks of Experiment 2, LiP-LLM still produced robust plans. As TABLE \ref{table:exp2_level} indicates, the comparative method’s success rate declined under stricter constraints, while LiP-LLM was less affected. The comparative method often ignored robot constraints, leading to errors, whereas LiP-LLM made no such errors.
        \item SPL: 
        Similar trends were observed in Experiment 1, with LiP-LLM performing best, followed by SMART-LLM and RoCo. The maximum difference from the baseline was 0.19. SPL highlighted task allocation efficiency, where the proposed method excelled with fewer redundant tasks.
        \item Process Time: 
        As shown in TABLE \ref{table:exp2_level}, unlike in Experiment 1, SMART-LLM had the shortest computation time. LiP-LLM took about 56 seconds longer than the baseline due to the increased number of API calls required for skill generation in long-term task planning.
    \end{enumerate}

\begin{table}[tb]
    \caption{\\Experiment 2: Results for each instruction level}
    \centering
    \renewcommand{\arraystretch}{1.2}
    
    \begin{tabularx}{\linewidth}{|p{0.4em}|D||C|C|C|}\hline
         & \multirow{2}{*}{methods} & \multicolumn{3}{|c|}{Environment C}\\ \cline{3-5}
         & ~ &Success Rate$\uparrow$& SPL $\uparrow$ & Time[s] $\downarrow$ \\\hline\hline
        \multirow{3}{*}{\hspace{-0.5em}L1} & LiP-LLM & \uuline{$\bm{0.58}$} & \uuline{$\bm{0.36}$} & \uline{72.9}\\
        ~ & RoCo & \uline{0.36} & 0.19 & 113 \\
        ~ & \hspace{-0.54em}SMART-LLM & 0.24 & \uline{0.24} & \uuline{$\bm{40.4}$} \\\hline\hline
        \multirow{3}{*}{\hspace{-0.5em}L2} & LiP-LLM & \uuline{$\bm{0.31}$} & \uuline{$\bm{0.16}$} & \uline{77.2}\\
        ~ & RoCo & \uline{0.18} & 0.13 & 137 \\
        ~ & \hspace{-0.54em}SMART-LLM & 0.16 & \uline{0.15} & \uuline{$\bm{35.1}$} \\\hline\hline
        \multirow{3}{*}{\hspace{-0.5em}L3} & LiP-LLM & \uuline{$\bm{0.31}$} & \uuline{$\bm{0.19}$} & \uline{96.9}\\
        ~ & RoCo & 0.00 & 0.00 & - \\
        ~ & \hspace{-0.54em}SMART-LLM & \uline{0.04} & \uline{0.04} & \uuline{$\bm{40.8}$} \\\hline
        
    \end{tabularx}
    \label{table:exp2_level}
\end{table}

\begin{table}[tb]
    \caption{\\Experiment 2: Result for each constraint}
    \centering
    \renewcommand{\arraystretch}{1.2}
    
    \begin{tabularx}{\linewidth}{|p{0.4em}|D||C|C|C|}\hline
         & \multirow{2}{*}{methods} & \multicolumn{3}{|c|}{Environment C}\\ \cline{3-5}
         & ~ &Success Rate$\uparrow$& SPL $\uparrow$ & Time[s] $\downarrow$   \\\hline\hline
        \multirow{3}{*}{\hspace{-0.5em}C1} & LiP-LLM & \uuline{$\bm{0.40}$} & \uuline{$\bm{0.24}$} & \uline{81.4}\\
        ~ & RoCo & \uline{0.24} & 0.15 & 122 \\
        ~ & \hspace{-0.54em}SMART-LLM & 0.22 & \uline{0.21} & \uuline{$\bm{39.0}$} \\\hline\hline
        \multirow{3}{*}{\hspace{-0.5em}C2} & LiP-LLM & \uuline{$\bm{0.38}$} & \uuline{$\bm{0.24}$} & \uline{81.1}\\
        ~ & RoCo & \uline{0.18} & \uline{0.09} & 141 \\
        ~ & \hspace{-0.54em}SMART-LLM & 0.09 & \uline{0.09} & \uuline{$\bm{33.4}$} \\\hline\hline
        \multirow{3}{*}{\hspace{-0.5em}C3} & LiP-LLM & \uuline{$\bm{0.42}$} & \uuline{$\bm{0.23}$} & \uline{78.2}\\
        ~ & RoCo & 0.11 & 0.08 & 85.5 \\
        ~ & \hspace{-0.54em}SMART-LLM & \uline{0.13} & \uline{0.13} & \uuline{$\bm{41.2}$} \\\hline
        
    \end{tabularx}
    \label{table:exp2_condition}
\end{table}

\section{Discussion}

\subsection{Effectiveness}

\subsubsection{Success Rate}
As detailed in Section IV, LiP-LLM showed high success rates in task planning across all conditions in Experiments 1 and 2, while comparative methods experienced significant drops in success rates with more robots or varying command types.
This result is attributed to the proposed method’s ability to mitigate LLM hallucinations by selecting skill lists through likelihood calculations and enforcing order constraints inferred from dependencies. In contrast, comparative methods failed due to command hallucinations and task failures from ignoring order constraints, such as simultaneous stacking tasks causing collisions. In addition, robustness to the increased number of robots contributed to success. In Experiment 2, with five robots and two rooms (kitchen and living room), accurate location tracking was essential, as shown by the success rate differences between Levels 2 and 3 in TABLE \ref{table:exp2_level}. Unlike the LLM-based method, which struggled with task allocation due to tracking limitations, LiP-LLM’s use of linear programming led to fewer allocation errors.
Regarding the scaling of the number of robots, theoretically, linear programming can be used to assign an optimal task to any number of robots. In these experiments, LiP-LLM demonstrated almost no failures in task allocation when using linear programming.

\subsubsection{Efficiency}
The SPL results indicate that LiP-LLM provides task planning results closer to the optimal steps set by humans. 
The proposed method considers dependencies among skills in the process of dependency graph generation, thereby enabling more tasks to be executed in parallel and more efficient task execution.

\subsubsection{Process Times}
In Experiment 1, LiP-LLM demonstrated reduced computation time for task planning.
The proposed method leverages combinatorial optimization for task allocation, resulting in much shorter process times compared to the comparative methods, which rely on time-intensive LLM inference. Reducing frequent LLM calls and using algorithmic approaches effectively shorten process time, though this was not observed in Experiment 2. The process time of LiP-LLM increases with task plan length, whereas SMART-LLM mitigates this by decomposing tasks into a single inference, improving process time. Although SMART-LLM exhibited advantages in process time, its performance was not sufﬁcient.
Additionally, the application of linear programming affects the scaling of the number of robots. Compared with RoCo, which is a decentralized system, LiP-LLM requires a shorter planning time. This result suggests that the proposed method is useful for scaling the number of units.

\subsection{Limitations}

\subsubsection{Challenges in Skill List Generation}
The creation of a predefined skill set is a challenge.
The skill list was generated using SayCan, which calculates the likelihood of each predefined skil from a prompt. A skill is defined by a function name and arguments, such as "pick\_and\_place(red block, blue block)". In environments with multiple objects, the number of predefined skills increasing owing to large number of object combinations.
In addition, it is not feasible to generate skills that are not predefined.
For example, if a door needs to be opened to move between rooms, the task would be impossible without a predefined "open door" skill.

\subsubsection{Challenges in Dependency Graph Generation}
In dependency graph generation, the increase in environmental information and diversification of tasks due to the complexity of the environment is a challenge.
In this experiment, the success rate of environment C was clearly lower than those of environments A and B. The lower success rate can be attributed to both increased environmental complexity and skill adjustments within the robot due to task diversification. As the number of rooms increases, the environmental information input to the LLM increases, complicating tasks such as moving objects between rooms. To address scalability issues, methods that can adequately handle the complexity introduced by environmental changes, including inter-robot skill coordination, must be developed.

\section{Conclusion}



In this study, we proposed a multi-robot task planning method LiP-LLM that integrates dependency graph generation using LLMs and task allocation via linear programming. By generating a dependency graph between skills, we demonstrated that skills with order constraints can be executed in the appropriate sequence, leading to more accurate and efficient task execution. Additionally, by adopting optimization methods for the task allocation step, we showed that the computation time can be reduced compared to methods that use LLMs for the same process.

Future studies should focus on applying LiP-LLM to real-world environments. The simulation environment used in this study was ideal, with known object coordinates. However, obtaining accurate environmental information in real environments is challenging and the available data may be limited. Therefore, it is necessary to improve the updating of environmental information through sequential environmental recognition and to enhance the methods for calculating weights used in linear programming.
In addition, the selected actions of a robot in a real environment may not always be successful. For example, a robot may fail to pick up or drop a red block. 
To handle such situations more effectively, it is crucial to develop a robust task planning method that includes sequential environmental recognition and task replanning 
with feedback, as demonstrated by Huang et al.\cite{huang2023inner}. This ensures that robots can adapt to dynamic environments and unexpected results, thereby improving overall task execution reliability.

\footnotesize
\bibliographystyle{junsrt}
\bibliography{main}

\end{document}